\pdfoutput=1

\documentclass[11pt]{article}
\usepackage[final]{acl} 

\usepackage{hyperref}
\usepackage{times}
\usepackage{latexsym}

\usepackage[T1]{fontenc}

\usepackage[utf8]{inputenc}
\usepackage{newunicodechar}
\newunicodechar{−}{\textminus}

\usepackage{CJKutf8}
\usepackage{CJK}

\usepackage{microtype}

\usepackage{inconsolata}

\usepackage{graphicx}

\usepackage{tabularx,array,booktabs}
\newcolumntype{Y}{>{\centering\arraybackslash}X} 

\usepackage{amsmath}

\title{Quality-Aware Translation Tagging in Multilingual RAG system}

\author{Hoyeon Moon\footnotemark[1]\\ Yonsei University \\ mhy9910@yonsei.ac.kr
        \And
        Byeolhee Kim\thanks{Equal contribution.} \\ University of Ulsan\\ College of Medicine \\ kbh0216@amc.seoul.kr
        \And
        Nikhil Verma\thanks{Corresponding author} \\ LG Electronics, Toronto AI Lab  \\ nikhil.verma@lge.com 
        }

\begin{document}

\maketitle

\begin{abstract}
Multilingual Retrieval-Augmented Generation (mRAG) often retrieves English documents and translates them into the query language for low-resource settings. 
However, poor translation quality degrades response generation performance. 
Existing approaches either assume sufficient translation quality or utilize the rewriting method, which introduces factual distortion and hallucinations. 
To mitigate these problems, we propose Quality-Aware Translation Tagging in mRAG (QTT-RAG), which explicitly evaluates translation quality along three dimensions-semantic equivalence, grammatical accuracy, and naturalness \& fluency-and attaches these scores as metadata without altering the original content. 
We evaluate QTT-RAG against CrossRAG and DKM-RAG as baselines in two open-domain QA benchmarks (\textsc{XORQA}, \textsc{MKQA}) using six instruction-tuned LLMs ranging from 2.4B to 14B parameters, covering two low-resource languages (Korean and Finnish) and one high-resource language (Chinese).
QTT-RAG outperforms the baselines by preserving factual integrity while enabling generator models to make informed decisions based on translation reliability. 
This approach allows for effective usage of cross-lingual documents in low-resource settings with limited native language documents, offering a practical and robust solution across multilingual domains. Code available at \url{https://github.com/HoyeonM/QTT-RAG}.
\end{abstract}

\section{Introduction}

\begin{figure}[t] 
  \centering
  \includegraphics[width=\linewidth]{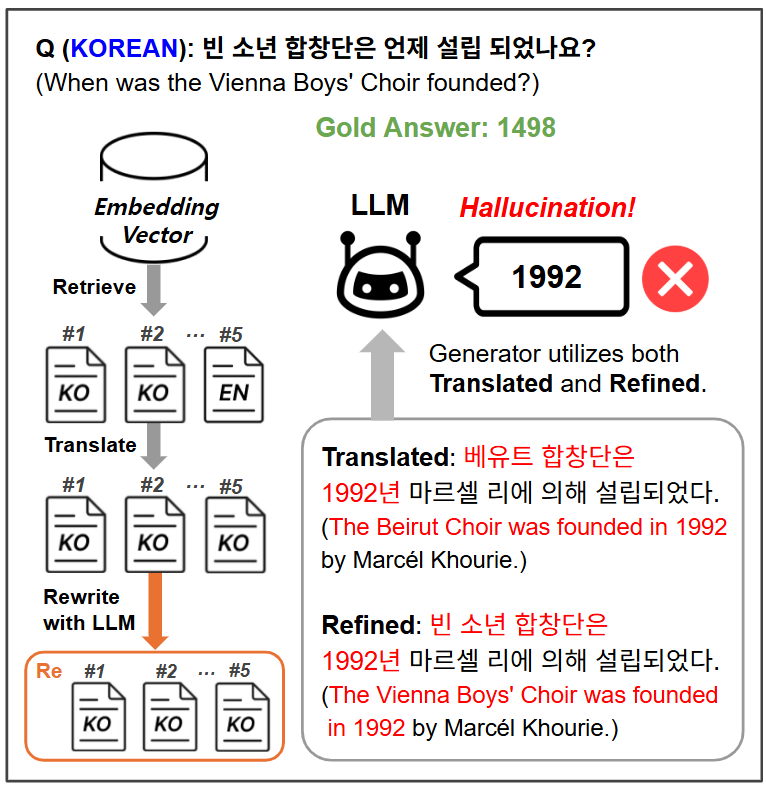}
  \caption{This figure illustrates a failure case of the previous approach (DKM-RAG). Hallucination arises when the LLM rewrites the translated documents, causing the generator to eventually produce an incorrect answer.}
  \label{fig:dkmrag_hallucination}
\end{figure}

\begin{figure*}[t]
    \centering
    \includegraphics[width=\textwidth]{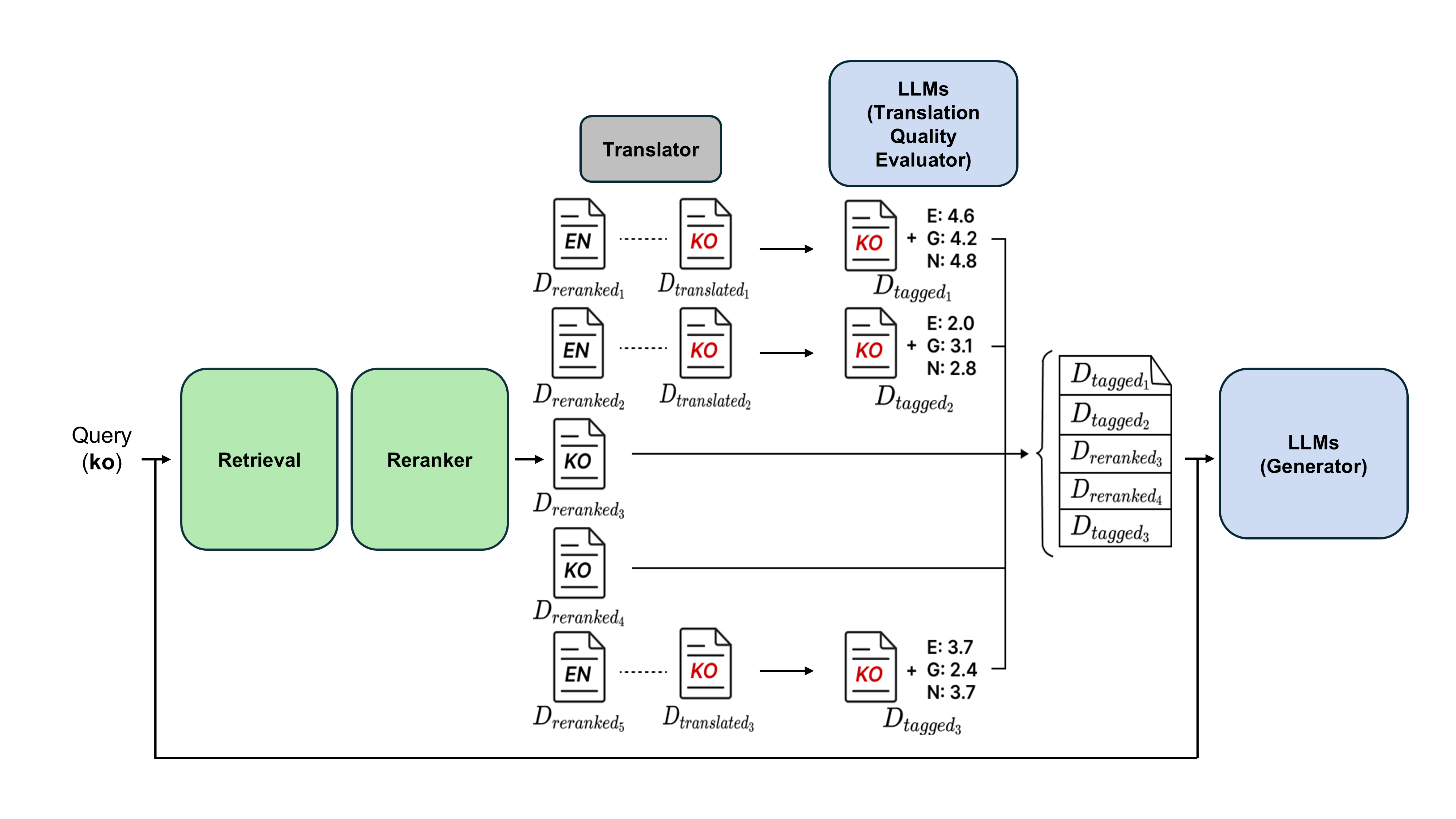}
    \caption{Overview of QTT-RAG System. After retrieving and reranking the top-5 most relevant documents, documents originally written in the query language (KO) are passed directly to the generator, whereas documents in foreign languages (EN) are translated and automatically scored along three dimensions: semantic equivalence (E), grammatical accuracy (G), and naturalness \& fluency (N). These passages are then re-inserted with the corresponding quality tags. The generator receives this quality-aware, tagged input, enabling it to produce factually grounded and translation-sensitive responses.}
    \label{fig:QTT_pipleline}
\end{figure*}

Retrieval-augmented generation (RAG) has become a standard approach for large language models (LLMs) for open-domain question answering tasks by accessing external sources of knowledge \citep{Lewis2020RAG}. 
One core challenge in multilingual RAG (mRAG) is retrieving relevant documents in a different language that would not degrade the quality of the generation.
This difficulty is exacerbated by a data imbalance: high-resource languages such as English dominate web-scale corpora, while medium- and low-resource languages (e.g., Korean, Finnish) remain underrepresented. 
This imbalance leads to inconsistent performance quality across languages in LLMs, even in safety and reliability issues~\citep{shen2024languagebarrierdissectingsafety}.

When queries and retrieved documents are in different languages, retrievers fail to identify relevant passages, and generators tend to produce code-switched or inaccurate responses \citep{ParkLee2025}. 
The same study also shows that performance improves substantially when the retrieved passages match the query language, highlighting a strong preference for the query language. 
This mismatch problem leads to a strong language preference bias, whereby generation performance improves when retrieved passages match the query language.
\citet{wang2024mitigatinglanguagemismatchrepetition} shows that LLM performance drops when input and output languages are mismatched, often leading to repetition and incoherence in multilingual generation and translation.

To address the language mismatch problem, two primary approaches have been explored: 1) Translating queries into English to match the dominant language of document collections, 2) Translating documents into the query language. 
Research in Cross-Lingual Information Retrieval (CLIR) has shown that document translation outperforms query translation \citep{Oard1999CLIR, SalehPecina2020, valentini2025cliruditcrosslingualinformationretrieval, yang2024translatedistilllearningcrosslanguagedense}. 
Recent work in mRAG has reinforced these findings, which shows that translating documents into the query language maintains cultural knowledge and word sense boundaries more accurately \citep{ParkLee2025}.

DKM-RAG \citep{ParkLee2025} introduces a document-centric approach that translates retrieved English passages into the query language and refines them using an LLM-based rewriting module.
Its refining method removes redundant sentences, ensures natural connections with the original text, and produces smooth query language writing. 
However, DKM-RAG has a key limitation: its refinement process can invoke hallucinations by inadvertently altering factual content, making irrelevant passages appear artificially relevant to the query, as shown in Figure~\ref{fig:dkmrag_hallucination}. 
It even refines the retrieved documents that are already in query language, unnecessarily modifying their contents and potentially distorting the original information.
Beyond these content-level issues, recent studies have revealed deeper limitations of LLMs in multilingual contexts, such as cultural commonsense understanding~\cite{shen2024understandingcapabilitieslimitationslarge}, as well as barriers in transferring knowledge across languages~\cite{chua2025crosslingualcapabilitiesknowledgebarriers}.

To address such problems, we propose Quality-Aware Translation Tagging in Multilingual RAG (QTT-RAG). 
Our approach employs explicit quality assessment instead of implicit quality control mechanisms. 
Specifically, we translate only those documents that are not already in the query language into the target language, and then employ an LLM to assess the translation quality based on three criteria: semantic equivalence, grammatical accuracy, and naturalness \& fluency.
Unlike implicit quality control approaches such as CrossRAG and DKM-RAG, which either assume adequate translation quality (CrossRAG) or rely on rewriting passages to improve fluency (DKM-RAG), our quality assessment method preserves factual integrity by providing detailed quality scores as metadata. This allows the generation model to make informed decisions without altering the original semantic content.

{Our key contributions are as follows:}
\begin{itemize}
    \item \textbf{LLM-based Translation Quality Assessment:} We propose an LLM-driven evaluation module that scores translation quality based on semantic equivalence, grammatical correctness, and linguistic naturalness.

    \item \textbf{QTT-RAG Architecture:} We introduce QTT-RAG, a multilingual RAG pipeline that attaches translation quality scores as metadata, enabling the generator to weigh information sources more reliably and thereby reducing factual distortion.

    \item \textbf{Empirical Validation:} Experiments across multilingual benchmarks show that QTT-RAG consistently improves 3-gram recall and robustness to translation errors compared to existing baselines such as CrossRAG and DKM-RAG.
\end{itemize}


\section{Background}
\subsection{Multilingual RAG}
Traditional Retrieval-Augmented Generation (RAG) systems primarily rely on English documents, retrieving and generating responses using dense passage encoders. Recent works have extended RAG to multilingual settings by integrating multilingual retrievers such as LaBSE \citep{Feng2022labse} and BGE-M3, often in combination with cross-lingual LLMs. However, \citet{Chirkova2024} demonstrate persistent language preference bias in multilingual RAG systems: generators achieve better performance when retrieved passages are in the same language as the query language but degrade when the context contains mixed or mismatched languages.

Two main strategies have been proposed to address the language mismatch: (i) query translation (tRAG), which translates the user query into English before retrieval, and (ii) document translation (CrossRAG), which translates all retrieved passages into a single language \citep{Ranaldi2025}. Query translation approaches suffer from information loss when relevant documents exist only in the original language, while document translation approaches may introduce translation noise that affects the generation stage. To improve document translation quality, DKM-RAG \citep{ParkLee2025} applies an LLM-based rewriting step to translated passages, enhancing fluency but at the risk of factual distortion.

Despite these advances and their notable contributions to mRAG,
existing methods still cannot reliably prevent translation-induced hallucinations. In contrast, our proposed QTT-RAG introduces an explicit quality evaluation framework that preserves the benefits of document translation while mitigating the risk of factual distortion. Rather than modifying content, QTT-RAG leverages quality assessments as metadata to better guide the generation process.

\section{Methodology}
We address a cross-lingual retrieval scenario where user queries $q$ are posed in medium or low-resource languages $L_q$ (e.g., Korean, Finnish), while the target document collection $\mathcal{D}$ predominantly contains documents in high-resource languages $L_h$ (e.g., English).

Our proposed pipeline, shown in Figure~\ref{fig:QTT_pipleline} consists of five sequential modules designed to handle cross-lingual retrieval and generation: (1) retrieval, (2) reranking, (3) language detection \& translation, (4) quality tagging, and (5) generation.

\subsection{Document Retrieval and Reranking}
Given a user query $q$ in language $L_q$, we first retrieve the top-$k$ candidate documents $D_k$ from the document collection $\mathcal{D}$.

For this initial retrieval step, we use BGE-M3, a state-of-the-art multilingual dense retrieval model that maps both queries and documents into a shared cross-lingual embedding space through a dual-encoder architecture.

The retrieved candidate list $D_k$ is then reranked using BGE-M3 as the reranking model, producing a reordered set of documents $D_{reranked}$. This reranking step computes more precise relevance scores for each query-document pair, enabling improved ranking of the initially retrieved candidates based on deeper semantic understanding.

\subsection{Cross-lingual Document Translation}
For each document $d \in D_{reranked}$, we first perform automatic language detection to identify its source language $L_d$. 

Documents already in the query language ($L_d = L_q$) bypass the translation process and are preserved in its original form, thereby avoiding unnecessary translation artifacts.
For documents in other languages ($L_d \neq L_q$), we employ neural machine translation using NLLB-200-600M, a multilingual translation model supporting over 200 languages. 
The model translates each document $d$ from its source language $L_d$ into the query language $L_q$, producing the translated document $D_{translated}$.

\subsection{Quality-Aware Translation Tagging}
\label{sec:qat_tagging}
We use an LLM-based agent to evaluate the translation quality of documents in $D_{translated}$ with the structured prompt shown in Table~\ref{tab:translation-prompt} of Appendix~\ref{sec:appendix_prompts}. 
The agent assesses each translated document across three criteria:

\begin{itemize}
\item \textbf{Semantic Equivalence}: Verifies that the translation faithfully preserves the original meaning and factual content.
\item \textbf{Grammatical Accuracy}: Evaluates syntactic, morphological, and structural correctness in the target language.
\item \textbf{Naturalness and Fluency}: Assesses whether the translation reads smoothly and idiomatically to native speakers.
\end{itemize}

Each criterion is scored based on the ELO rating system from 0.0 to 5.0. We attach these quality scores as tags to each translated document, creating the quality-tagged document $D_{tagged}$. 
Examples of the tagged documents can be found in Table~\ref{tab:translation_quality_case_low} and Table~\ref{tab:translation_quality_case_high} in Appendix \ref{sec:appendix_translation_quality}, where Table~\ref{tab:translation_quality_case_low} represents low-quality translation cases and Table~\ref{tab:translation_quality_case_high} shows high-quality translation cases for Korean, Finnish, and Chinese.
If a document is originally written in the query language, no quality score is added.
This tagging approach preserves and fully utilizes all translated documents while providing the quality information to guide the generation model.

\subsection{Response Generation}
The generator LLM receives the user query $q$ concatenated with the quality-tagged document set through a structured prompt template detailed in Table~\ref{tab:generation-prompt} of Appendix~\ref{sec:appendix_prompts}.
Rather than employing additional fine-tuning, we leverage in-context learning by explicitly exposing the quality scores within the input prompt.  

The template instructs the LLM to prioritize passages with higher quality scores, enabling responses to rely more heavily on high-quality translations while down-weighting or cautiously handling lower-quality passages.

\section{Experiments and Results}
In this section, we describe the datasets used in our experiments, the experimental setup, evaluation metrics, and results, followed by ablation studies to analyze the contribution of each component.

As baselines, we compare our method against three approaches: (i) Base, a retrieval-only system without translation, which relies solely on reranked retrieved documents; (ii) CrossRAG, which translates all retrieved passages into the query language; and (iii) DKM-RAG, which refines translated passages using an LLM-based rewriting step.

\subsection{Dataset}
We conduct experiments on two multilingual open-domain QA benchmarks: MKQA: Multilingual Knowledge Questions \& Answers~\citep{Longpre2020mkqa} and XOR-TyDi: Cross-lingual Open-Retrieval Question Answering~\citep{Asai2021xortydiqa} datasets for multilingual open-domain question answering tasks. 
MKQA consists of 10,000 examples from the Natural Questions (NQ) benchmark \citep{Kwiatkowski2019nq}, translated into 26 languages, creating parallel multilingual QA pairs grounded in English Wikipedia. 
However, MKQA does not provide document-level annotations. For consistency with prior benchmarks that include gold document labels, we therefore adopt a subset of 2,827 MKQA samples that overlap with KILT-NQ (Knowledge Intensive Language Tasks Natural Questions).
%

XOR-TyDi QA extends the TyDi QA \citep{Clark2020TydiQA} benchmark by introducing cross-lingual open retrieval challenges, where questions are written in typologically diverse languages and paired with English Wikipedia articles. 

In our experiments, we use the Korean, Finnish, and Chinese splits of MKQA. For XOR-TyDi QA, we evaluate on the Korean and Finnish splits, comprising 371 and 615 questions respectively.

\subsection{Experimental Setup}
We implement our QTT-RAG framework using Bergen \citep{Rau2024BERGEN} as the experimental framework and conduct baseline comparisons on Korean, Finnish, and Chinese language settings. 

\paragraph{Knowledge Base}\hspace{0.5em}
We construct our document index from Wikipedia, comprising 25M English, 1.6M Korean, 1.5M Finnish, and 11M Chinese examples. 
Wikipedia is selected for two main reasons: (i) both XOR-TyDi QA and MKQA are curated against Wikipedia pages, ensuring high answer coverage; (ii) it offers broad multilingual coverage with consistent article quality and structured formatting across languages.

\paragraph{Retrieval \& Reranking}\hspace{0.5em}
We adopt a two-stage retrieval pipeline: (i) an initial dense retriever to maximize recall over a large index, (ii) followed by a reranker that re-scores the top-$K$ candidates through query–passage interactions to improve precision at early ranks. 
This is crucial because only a limited number of passages can be provided to the LLM. Reranking ensures that answer-bearing passages are prioritized while topical but non-answer passages and near duplicates are suppressed.

We choose BGE-M3~\citep{Xiao2024bgem3} as both retriever and reranker for three practical reasons: (i) it provides a single multilingual checkpoint with strong cross-lingual retrieval across 100+ languages; (ii) it has been adopted in prior work such as DKM-RAG and CrossRAG, enabling direct comparability; and (iii) it offers publicly available weights and a built-in reranker, facilitating reproducibility.

\paragraph{Translation}\hspace{0.5em}
Documents that are retrieved in languages other than the query language are translated by NLLB-200-distilled-600M (NLLB) \citep{Costa-jussa2022nllb}, a multilingual neural machine translation model supporting more than 200 languages. 
NLLB achieves BLEU scores in the 30–40 range for many low-resource language pairs, making it a strong baseline for translation quality. 
While NLLB offers credible and scalable translation capabilities, relying solely on translated content can still introduce errors or stylistic inconsistencies. 
This limitation motivates our design choice to incorporate translation quality assessment, allowing the generator to dynamically weigh the reliability of translated passages rather than treating all translations equally.

\paragraph{Translation Quality Assessment}\hspace{0.5em}
We adapt Llama-3.1-8B-Instruct \citep{Dubey2024llama3} as our quality assessment agent to evaluate translation quality across three criteria (semantic equivalence, grammatical accuracy, and naturalness \& fluency) as described in Section~\ref{sec:qat_tagging}. For each query language, we design the assessment prompt in the same language. The exact prompts are provided in Table~\ref{tab:generation-prompt} of Appendix~\ref{sec:appendix_prompts}.

\paragraph{Response Generation}\hspace{0.5em}
We evaluate our framework with six pretrained, instruction-tuned language models of varying scales: Exaone-3.5-2.4B-Instruct, Exaone-3.5-7.8B-Instruct \citep{Yoo2024exaone}, Qwen2.5-7B-Instruct \citep{Yang2024qwen25}, Llama-3.1-8B-Instruct \citep{Dubey2024llama3}, Aya-Expanse-8B \citep{Aryabumi2024aya}, and Phi-4 (14b) \citep{Abdin2024phi4}. 
This diverse set of models enables us to assess the generalization ability of our approach across different model architectures and capabilities. 

\paragraph{Evaluation Metric}\hspace{0.5em}
We use character 3-gram recall as the evaluation metric~\cite{Chirkova2024}. Given a gold answer, character 3-gram recall first extracts all overlapping three-character sequences (trigrams) from the entire gold string. The score is then calculated as the proportion of these gold trigrams that appear anywhere in the model's prediction. Character 3-gram recall is well-suited for multilingual QA as it tolerates orthographic variations while still penalizing hallucinations and missing content. Unlike word-level metrics, this character-level approach is language-agnostic and requires no language-specific processing, making it well-suited for cross-lingual evaluation.

\subsection{Failure Cases of DKM-RAG and CrossRAG}
DKM-RAG improves translation quality by rewriting retrieved passages conditioned on the query. Although this process can mitigate noisy translations, it often results in knowledge drift, where the rewritten passages introduce query terms or assert relations unsupported by the original documents.
To validate this, we manually analyze 1{,}855 retrieved documents for 371 questions from XOR-TyDi--ko. In 214 cases (11.5\%), entities in the query (e.g. names, places, and dates) that were absent from the original documents are added during rewriting.
This rate of entity hallucination indicates a notable limitation of rewriting-based approaches.

Table~\ref{tab:dkm_hallucination_cases} illustrates how rewriting can change the factual content. 
In Case 1, the retrieved passage describes an unrelated person named “Rumer Godden”. However the rewritten output asserts a death date for “Gwisil Boksin,” bridging the query to irrelevant evidence and fabricating a fact that the source does not contain. The downstream generator then treats the rewritten passage as authoritative and produces the fabricated answer.

Table~\ref{tab:crossrag_wrongtrans_cases} presents a failure case of translation in the CrossRAG method. In this case, the original retrieved passage is incorrectly translated, omitting an important part of the original content.

\begin{table}[h]
\centering\small
\begin{tabularx}{\columnwidth}{lX}
\toprule
\multicolumn{2}{l}{\textbf{Case 1}} \\
\midrule
\textbf{Query} & \begin{CJK}{UTF8}{mj}귀실복신 사망일은 언제인가요?\end{CJK} (When did Gwisil Boksin die?) \\
\textbf{Retrieved} & Rumer Godden died on 8 November 1998, aged 90, following a stroke... \\
\textbf{Refined} & \begin{CJK}{UTF8}{mj}귀실복신의 사망일은 1998년 11월 8일입니다.\end{CJK} (Gwisil Boksin's date of death is November 8, 1998.) \\
\textbf{LLM Output} & \begin{CJK}{UTF8}{mj}주어진 정보만으로는 1998년 11월 8일이 가장 유력한 답변입니다.\end{CJK} (Based on the given information, November 8, 1998 is the most likely answer.) \\
\bottomrule
\end{tabularx}
\caption{Case study of factual distortion in DKM-RAG for a Korean query.}
\label{tab:dkm_hallucination_cases}
\end{table}

\begin{table}[h]
\centering\small
\begin{tabularx}{\columnwidth}{lX}
\toprule
\multicolumn{2}{l}{\textbf{Case 2}} \\
\midrule
\textbf{Query} & \begin{CJK}{UTF8}{mj}북유럽의 노르딕 국가는 몇개인가요?\end{CJK} (How many Nordic countries are there in Northern Europe?) \\
\textbf{Retrieved} & "Scandinavia" is sometimes used as a synonym for the Nordic countries, although within the Nordic countries the terms are considered distinct. \\
\textbf{Translated} & \begin{CJK}{UTF8}{mj}북유럽 국가들 내에서는 스칸디나비아라는 용어가 구별되는 것으로 간주된다.\end{CJK} (Within the Nordic countries, the term Scandinavia is regarded as distinct.) \\
\bottomrule
\end{tabularx}
\caption{Case study of incorrect translation in CrossRAG for a Korean query.}
\label{tab:crossrag_wrongtrans_cases}
\end{table}

\subsection{Quality-Aware Translation Tagging}
Our QTT-RAG explicitly tags translation quality as metadata using an LLM without rewriting retrieved content.
Unlike refinement-based methods, which risk distorting original information into inaccurate content, our approach preserves the original translations and supplements them with quality scores as metadata.
This non-destructive design enables the generation model to prioritize higher-quality sources while maintaining access to potentially useful information from lower-quality translations. We validate this advantage through experiments across three languages—Korean, Finnish, and Chinese—where QTT-RAG consistently outperforms baseline methods.

\paragraph{Korean}\hspace{0.5em}
Korean is considered a low-resource language~\cite{jang2024kodialogbenchevaluatingconversationalunderstanding}. As shown in Table~\ref{tab:main_result}, QTT-RAG consistently outperforms all baselines on \textsc{XOR-TyDi}--ko and \textsc{MKQA}--ko across six LLMs.
In Korean, performance gains range from 0.4\% to 6.8\% over the baselines. 
Among the evaluated models, Exaone-3.5-7.8B-Instruct achieves the highest score, which is expected given its training on a collection of instruction-tuned bilingual (English–Korean) generative models.

\begin{table}[t]
\centering\small
\begin{tabularx}{\linewidth}{lYYYY}
\toprule
& \multicolumn{4}{c}{\textbf{Character 3-gram Recall (\%)}}\\
\cmidrule(lr){2-5}
\textbf{Model} & Base & Cross & DKM & QTT\\
\midrule
\multicolumn{5}{c}{\textbf{XOR-TyDi--ko}}\\
\midrule
Exaone-3.5-2.4B-Instruct   & 37.0 & 37.3 & 35.1 & \textbf{41.3}\\
Qwen2.5-7B-Instruct    & 34.3 & 36.5 & 34.2 & \textbf{36.9}\\
Exaone-3.5-7.8B-Instruct   & 40.7 & 42.0 & 39.7 & \textbf{43.8}\\
Aya-Expanse-8B        & 38.2 & 39.7 & 37.0 & \textbf{42.8}\\
Llama-3.1-8B-Instruct   & 33.7 & 34.2 & 33.7 & \textbf{37.2}\\
Phi-4 (14B)       & 40.6 & 41.0 & 35.7 & \textbf{42.5}\\
\midrule
\multicolumn{5}{c}{\textbf{MKQA--ko}}\\
\midrule
Exaone-3.5-2.4B-Instruct   & 29.2 & 30.1 & 32.0 & \textbf{36.0}\\
Qwen2.5-7B-Instruct    & 28.6 & 28.5 & 30.6 & \textbf{33.3}\\
Exaone-3.5-7.8B-Instruct   & 33.4 & 33.4 & 36.4 & \textbf{40.0}\\
Aya-Expanse-8B        & 32.6 & 33.8 & 35.5 & \textbf{39.0}\\
Llama-3.1-8B-Instruct   & 28.5 & 27.5 & 28.3 & \textbf{33.4}\\
Phi-4 (14B)       & 33.8 & 33.4 & 35.8 & \textbf{37.7}\\
\bottomrule
\end{tabularx}
\caption{Character 3-gram recall (\%) on the \textsc{XOR-TyDi} and \textsc{MKQA} benchmarks (Korean subset). 
Six LLMs are evaluated under four retrieval pipelines:
\textbf{Base}, \textbf{Cross} = CrossRAG,
\textbf{DKM} = DKM-RAG, and \textbf{QTT} = QTT-RAG.}
\label{tab:main_result}
\end{table}

\paragraph{Finnish}\hspace{0.5em}
Finnish is also considered a low-resource language like Korean~\cite{ouzerrout-2025-uter}. 
Our method achieves comparable performance on the XOR-TyDi Finnish dataset except for one LLM. 
The results are shown in Table~\ref{tab:finnish_result}. 

\paragraph{Chinese}\hspace{0.5em}
Chinese is a high-resource language~\cite{jang2024kodialogbenchevaluatingconversationalunderstanding}, which most of the top-ranked passages are already in Chinese. As a result, opportunities for cross-lingual translation are limited, leaving less headroom for further gains. In the MKQA--zh experiment results (Table~\ref{tab:mkqa_zh_result}), CrossRAG achieves better performance with Exaone-3.5-2.4B-Instruct, Exaone-3.5-7.8B-Instruct, and Llama-3.1-8B-Instruct.

However, when a non-Chinese document appears, QTT-RAG’s explicit, non-rewriting quality cues benefit models that reliably follow metadata, resulting clear improvements with Aya-Expanse-8B, Qwen2.5-7B-Instruct, and Phi-4 (14B).

\begin{table}[t]
\centering\small
\begin{tabularx}{\linewidth}{l|YYYY}
\toprule
& \multicolumn{4}{c}{\textbf{Character 3-gram Recall (\%)}}\\
\cmidrule(lr){2-5}
\textbf{Model} & Base & Cross & DKM & QTT\\
\midrule
\multicolumn{5}{c}{\textbf{XOR-TyDi--fi}}\\
\midrule
Exaone-3.5-2.4B-Instruct   & 45.0 & 45.6 & \textbf{50.4} & \textbf{50.4}\\
Qwen2.5-7B-Instruct    & 55.9 & 56.7 & 55.7 & \textbf{58.6}\\
Exaone-3.5-7.8B-Instruct   & 56.0 & 55.6 & 56.1 & \textbf{59.3}\\
Aya-Expanse-8B        & 57.6 & \textbf{60.1} & 58.3 & 55.4\\
Llama-3.1-8B-Instruct   & 54.9 & 54.9 & 52.7 & \textbf{60.0}\\
Phi-4 (14B)       & 64.0 & 63.5 & 60.1 & \textbf{66.8}\\
\bottomrule
\end{tabularx}
\caption{Character 3-gram recall (\%) on the \textsc{XOR-TyDi}
benchmarks (Finnish subset). Six LLMs are evaluated under four retrieval pipelines:
\textbf{Base}, \textbf{Cross} = CrossRAG,
\textbf{DKM} = DKM-RAG, and \textbf{QTT} = QTT-RAG.}
\label{tab:finnish_result}
\end{table}

\begin{table}[t]
\centering\small
\begin{tabularx}{\linewidth}{l|YYYY}
\toprule
& \multicolumn{4}{c}{\textbf{Character 3-gram Recall (\%)}}\\
\cmidrule(lr){2-5}
\textbf{Model} & Base & Cross & DKM & QTT \\
\midrule
\multicolumn{5}{c}{\textbf{MKQA--zh}}\\
\midrule
Exaone-3.5-2.4B-Instruct         & 19.0  & \textbf{25.2}  & 23.9  & 24.4 \\
Qwen2.5-7B-Instruct & 27.7  & 30.0  & 28.7  & \textbf{31.9} \\
Exaone-3.5-7.8B-Instruct         & 22.2  & \textbf{26.2}  & 26.1  & 25.8 \\
Aya-Expanse-8B              & 26.3  & 32.8  & 33.2  & \textbf{33.9} \\
Llama-3.1-8B-Instruct& 25.2  & \textbf{30.1}  & 28.8  & 29.3 \\
Phi-4 (14B)             & 30.9  & 33.8  & 33.0  & \textbf{34.5} \\
\bottomrule
\end{tabularx}
\caption{Character 3-gram recall (\%) on the \textsc{MKQA} benchmark (Chinese subset). Six LLMs are evaluated under four retrieval pipelines:
\textbf{Base}, \textbf{Cross} = CrossRAG,
\textbf{DKM} = DKM-RAG, and \textbf{QTT} = QTT-RAG.}
\label{tab:mkqa_zh_result}
\end{table}

\begin{table}[t]
\centering\small
\setlength{\tabcolsep}{4pt}
\begin{tabularx}{\linewidth}{l|YY|YY}
\toprule
& \multicolumn{2}{c|}{\textbf{XOR-TyDi--ko}} & \multicolumn{2}{c}{\textbf{MKQA--ko}} \\
\cmidrule(lr){2-3}\cmidrule(lr){4-5}
\textbf{Model} & Hard & QTT & Hard & QTT\\
\midrule
Exaone-3.5-2.4B-Instruct  & 40.3 & \textbf{41.3} & 32.1 & \textbf{36.0} \\
Qwen2.5-7B-Instruct   & 36.6 & \textbf{36.9} & 30.6  & \textbf{33.3} \\
Exaone-3.5-7.8B-Instruct  & 43.2 & \textbf{43.8} & 34.2 & \textbf{40.0} \\
Aya-Expanse-8B       & 40.4 & \textbf{42.8} & 35.7 & \textbf{39.0} \\
Llama-3.1-8B-Instruct  & 35.0 & \textbf{37.2} & 28.7  & \textbf{33.4} \\
Phi-4 (14B)      & 40.2 & \textbf{42.5} & 33.7 & \textbf{37.7} \\
\bottomrule
\end{tabularx}
\caption{Ablation on filtering strategy. \textbf{Hard} = Hard filtering;
\textbf{QTT} = QTT-RAG. Values are Character 3-gram Recall (\%).}
\label{tab:ablation_hard_ko}
\end{table}

\begin{table}[t]
\centering\small
\begin{tabularx}{\linewidth}{l|YY|YY}
\toprule
& \multicolumn{2}{c|}{\textbf{XOR-TyDi--fi}} & \multicolumn{2}{c}{\textbf{MKQA--zh}} \\
\cmidrule(lr){2-3}\cmidrule(lr){4-5}
\textbf{Model} & Hard & QTT & Hard & QTT \\
\midrule
Exaone-3.5-2.4B-Instruct         & \textbf{51.3} & 50.4 & \textbf{25.4} & 24.4 \\
Qwen2.5-7B-Instruct & \textbf{58.7} & 58.6 & 30.1 & \textbf{31.9} \\
Exaone-3.5-7.8B-Instruct         & 58.7 & \textbf{59.3} & \textbf{26.8} & 25.8 \\
Aya-Expanse-8B              & \textbf{61.0} & 55.4 & 33.0 & \textbf{33.9} \\
Llama-3.1-8B-Instruct  & \textbf{60.0} & \textbf{60.0} & \textbf{29.9} & 29.3 \\
Phi-4 (14B)             & \textbf{66.8} & \textbf{66.8} & 33.7 & \textbf{34.5} \\
\bottomrule
\end{tabularx}
\caption{Ablation on filtering strategy. \textbf{Hard} = Hard filtering; \textbf{QTT} = QTT-RAG. Values are Character 3-gram Recall (\%).}
\label{tab:ablation_hard_fi_zh}
\end{table}

\subsection{Leveraging Translation Quality}
To examine our design choice of quality tagging, we conduct an ablation study comparing two strategies: 
(1) \textbf{Hard filtering}, which excludes documents that are below all specified quality thresholds; and 
(2) \textbf{QTT-RAG}, which is our proposed method utilizing quality scores as metadata.

For Hard filtering, we use the same prompt employed for translation quality evaluation (Table~\ref{tab:translation-prompt}) to obtain scores along three criteria: Semantic Equivalence, Grammatical Accuracy, and Naturalness \& Fluency.  
Based on these scores, we exclude documents if they fall below a threshold of 3.5 on all criteria.

Table~\ref{tab:ablation_hard_ko} and Table~\ref{tab:ablation_hard_fi_zh} show the comparison between Hard filtering and QTT-RAG. In Korean (Table~\ref{tab:ablation_hard_ko}), QTT-RAG consistently outperforms Hard filtering on all models, with average relative gains of 3.8\% on XOR-TyDi--ko and 12.6\% on MKQA--ko.
In XOR-TyDi--fi (Table~\ref{tab:ablation_hard_fi_zh}, left), the results are generally comparable across methods. Notably, Hard filtering achieves the best score on Aya-8B, outperforming QTT-RAG as well as all other baselines (Base, CrossRAG, and DKM-RAG).
In MKQA--zh (Table~\ref{tab:ablation_hard_fi_zh}, right), Hard filtering surpasses both QTT-RAG and CrossRAG on Exaone-3.5-2.4B-Instruct and Exaone-3.5-7.8B-Instruct. QTT-RAG remains the best on Aya-Expanse-8B, Phi-4 (14B), and Qwen2.5-7B-Instruct, while CrossRAG leads with Llama-3.1-8B-Instruct by a small performance difference.

With these additional experiments, we observe that effectiveness varies across languages and setups--such as resource level, the proportion of cross-lingual passages, retriever and MT quality, filtering thresholds, retained ratio, and the generator backbone--so no single strategy dominates universally. 


Hard filtering simplifies the generator input and can be effective in certain regimes, particularly when in-language evidence is already abundant, and removing a small set of low-scored translated passages leaves most relevant evidence intact. 
However, it risks discarding rare but critical information and is sensitive to choice of threshold and language. In contrast, QTT-RAG avoids brittle thresholds and preserves coverage, which is crucial when high-quality translations are sparse or unevenly distributed. 

Together, these findings suggest that while Hard filtering may offer gains under favorable conditions, quality tagging delivers more consistent improvements across languages and models.

\section{Discussion}
We analyze cases where QTT-RAG delivers smaller gains in Chinese compared to Korean and Finnish.  
To formalize this observation, we denote the cross-lingual share by
\[
r_{\text{lang}} \;=\; \frac{N_{\mathrm{translated}}}{N_{\mathrm{input}}}
\]

where $N_{\mathrm{translated}}$ denotes the number of translated documents and $N_{\mathrm{input}}$ denotes the total number of retrieved documents.

In our experiments, the MKQA--zh split has a relatively low cross-lingual share (\(r_{\text{lang}}=5.0\%\)), whereas the MKQA--ko split shows a much higher cross-lingual share (\(r_{\text{lang}}=22.7\%\)). This disparity helps explain why QTT-RAG’s improvements tend to be smaller in Chinese than in Korean: there are simply fewer instances where translated evidence is involved. More broadly, overall effectiveness also depends on retriever and MT quality, generator backbone, the distribution of retrieved languages, and the evaluation setting.

For future work, we aim to expand our evaluation to a wider set of languages to further test the scalability of our approach. We also plan to explore hybrid retrieval strategies, such as deliberately inducing cross-lingual usage via English-only retrieval for non-English queries.

\section{Conclusion}
We propose QTT-RAG, a novel multilingual RAG framework that introduces translation quality tagging as an explicit mechanism to mitigate factual distortions and translation-induced errors.
Unlike prior approaches such as CrossRAG, which assumes adequate translation quality, or DKM-RAG, which relies on rewriting and risk semantic drift, our method preserves the original translated content and supplements it with fine-grained quality scores as metadata.
Through extensive experiments on two multilingual QA benchmarks (XOR–TyDi QA and MKQA) across three typologically diverse languages—Korean, Finnish, and Chinese—and six instruction-tuned LLMs ranging from 2.4B to 14B parameters, we demonstrate that QTT-RAG consistently improves character 3-gram recall over strong baselines particularly in low-resource settings (Korean and Finnish).
Ablation studies further reveal that quality tagging offers a more reliable default than Hard filtering, while still leaving room for filtering-based strategies in specific regimes with abundant in-language evidence.

\section*{Limitations}

QTT-RAG is most effective when a substantial portion of retrieved documents is in a different language from the query and translation quality is heterogeneous.
In other words, when the majority of retrieved passages already match the query language, opportunities for translation and tagging diminish, and gains naturally become smaller.

One other limitation is that the generator must reliably interpret and utilize the structured metadata; models with weaker instruction-following capabilities or shorter effective context windows may fail to fully exploit these quality cues.

We also acknowledge that the experiments are limited to only few languages—Korean, Finnish and Chinese—which may be insufficient to generalize the effectiveness of our method. Further experiments on a more diverse set of languages are required to validate its broader applicability.

\section*{Acknowledgments}
We would like to thank Manasa Bharadwaj and Kevin Ferreira for their contributions and support throughout this project. This work was made possible by the support from LG Toronto AI Lab and CARTE, and we sincerely appreciate the opportunity to collaborate with them. This work was supported by the Institute of Information \& Communications Technology Planning \& Evaluation (IITP) grant funded by the Korea government (MSIT) (RS-2022-00143911, AI Excellence Global Innovative Leader Education Program). Byeolhee Kim was supported by a grant of the Korea Health Technology R\&D Project through the Korea Health Industry Development Institute (KHIDI), funded by the Ministry of Health \& Welfare, Republic of Korea (grant number: HR21C0198).

\newpage

\bibliography{main}

\newpage
\appendix

\section{Prompt Templates}
\label{sec:appendix_prompts}
Table~\ref{tab:translation-prompt} presents the template used for the LLM to assign translation quality scores for each translated document. It evaluates scores across three dimensions: Semantic Equivalence, Grammatical Accuracy, and Naturalness \& Fluency.

Table~\ref{tab:generation-prompt} shows the generation prompt template, which contains both System and User messages in three languages. This template instructs the LLM to prioritize passages with higher quality scores across all three dimensions, enabling generator to output quality-aware answer leveraging the most reliable translated contents first.

\begin{table*}[t]
\centering
\begin{tabular}{|p{0.95\linewidth}|}
\hline
\textbf{Translation Quality Assessment Prompt (Korean / Finnish / Chinese)} \\
\hline
\textbf{\textcolor{blue}{Korean}: } \\
\begin{CJK}{UTF8}{mj}영어 원문: \end{CJK} \{original english passage\} \\
\begin{CJK}{UTF8}{mj}한국어 번역문: \end{CJK} \{translated korean passage\} \\
\begin{CJK}{UTF8}{mj}다음 영어 원문과 한국어 번역문의 품질을 세 가지 기준(의미론적 일치성, 문법적 정확성, 자연스러움과 유창성)에 대해 각각 0.0점에서 5.0점 사이의 소수점 첫째 자리까지의 점수로 평가해주세요. 다른 설명 없이 JSON 형식으로만 응답해주세요. \end{CJK} \\[0.5ex]
\begin{CJK}{UTF8}{mj}\textbf{예시}:{"의미론적 일치성": 5.0, "문법적 정확성": 2.5, "자연스러움과 유창성": 4.3}\end{CJK} \\
\\[0.5em]
\textbf{\textcolor{blue}{Finnish}: } \\
Alkuperäinen teksti (englanti): \{original english passage\} \\
Käännös (suomi): \{translated finnish passage\} \\
Arvioi käännöksen laatu englanninkielisen alkuperäistekstin ja suomenkielisen käännöksen välillä kolmen kriteerin perusteella: semanttinen johdonmukaisuus, kieliopillinen tarkkuus ja luontevuus ja sujuvuus. Anna pisteet jokaiselle kriteerille välillä 0.0–5.0 yhdellä desimaalilla. Vastaa vain JSON-muodossa ilman mitään lisäselityksiä tai kommentteja. \\
\textbf{Esimerkki}: {"Semanttinen johdonmukaisuus": 5.0, "Kieliopillinen tarkkuus": 2.5, "Luontevuus ja sujuvuus": 4.3} \\
\\[0.5em]
\textbf{\textcolor{blue}{Chinese}: } \\
\begin{CJK}{UTF8}{gbsn}原文(英文): \{original english passage\} \end{CJK}\\
\begin{CJK}{UTF8}{gbsn}翻译(中文): \{translated chinese passage\} \end{CJK}\\
\begin{CJK}{UTF8}{gbsn}请根据以下三个标准评估英文原文与其中文翻译之间的翻译质量: 语义一致性、语法准确性、以及语言的自然流畅度。请为每个标准打分，分数范围为0.0到5.0，保留一位小数。只需以JSON格式作答，不要添加任何额外说明或评论。\end{CJK} \\
\begin{CJK}{UTF8}{gbsn} \textbf{示例}：{"语义一致性": 5.0, "语法准确性": 2.5, "语言流畅度": 4.3} \end{CJK} \\
\\[0.5em]
\textbf{English Version:} \\
Original Passage: \{original english passage\} \\
Translated Passage: \{translated \{\textcolor{blue}{query language}\} passage\} \\

Please evaluate the quality of the following English-to-\{\textcolor{blue}{query language}\} translation using the three criteria: Semantic Equivalence, Grammatical Accuracy and Naturalness \& Fluency from 0.0 to 5.0. Respond strictly in JSON format, without additional explanations. \\
\textbf{Example:}
"Semantic Equivalence": 5.0, "Grammatical Accuracy": 2.5, "Naturalness \& Fluency": 4.3 \\

\hline
\end{tabular}
\caption{The prompt used for evaluating translated passages based on three dimensions of translation quality. An example (few-shot) output format is also provided for better generation. Quality scores are then attached as metadata to each translated document.}
\label{tab:translation-prompt}
\end{table*}

\begin{table*}[t]
\centering
\begin{tabular}{|p{0.95\linewidth}|}
\hline
\textbf{Generation Prompt (Korean / Finnish / Chinese)} \\
\hline
\textbf{System (\textcolor{blue}{Korean}):} \begin{CJK}{UTF8}{mj}이제부터 너는 내 유능한 비서야. 내가 제공하는 문서들은 일부는 원래 한국어로 작성된 문서이고, 일부는 영어 원문을 한국어로 번역한 후 품질 평가 점수가 부여된 문서야. 번역된 문서에는 의미론적 일치성, 문법적 정확성, 자연스러움과 유창성에 대한 점수가 포함되어 있으며, 각각 0.0에서 5.0 사이의 값이야. 원래 한국어로 작성된 문서를 가장 신뢰하고 우선적으로 참고해 줘. 번역된 문서는 점수가 높은 순서대로 활용해 줘. 확신이 들지 않는 정보는 신중하게 판단해. 모든 질문에는 가능한 한 짧고 정확하게, 반드시 한국어로 대답해 줘.\end{CJK} \\[0.5ex]
\\[0.3em]

\textbf{System (\textcolor{blue}{Finnish}):} \begin{CJK}{UTF8}Olet nyt minun osaava assistenttini. Antamani asiakirjat ovat joko alun perin suomeksi kirjoitettuja tai englanninkielisestä alkuperästä suomeksi käännettyjä, ja niihin on liitetty laadun arviointipisteet. Käännetyillä asiakirjoilla on pistemäärät semanttisesta yhteneväisyydestä, kieliopillisesta oikeellisuudesta sekä luonnollisuudesta ja sujuvuudesta, asteikolla 0.0–5.0. Luota eniten alun perin suomeksi kirjoitettuihin asiakirjoihin ja käytä niitä ensisijaisesti. Käännösasiakirjoja voit käyttää apuna korkeimman pistemäärän mukaisessa järjestyksessä. Ole varovainen, jos tieto ei vaikuta varmalta. Vastaa kaikkiin kysymyksiin mahdollisimman lyhyesti ja tarkasti, aina suomeksi.{mj}\end{CJK} \\[0.5ex]
\\[0.3em]

\textbf{System (\textcolor{blue}{Chinese}):} \begin{CJK}{UTF8}{gbsn}你现在是我聪明能干的助手。我提供的文档有些是原始中文写成的，有些是从英文翻译成中文并附有质量评分的翻译文档。翻译文档包含三个评分指标：语义一致性、语法准确性和语言流畅度，评分范围为0.0到5.0。请优先参考原始中文文档，因为它们最可靠。翻译文档可以作为补充信息，按评分高低依次参考。对于不确定的信息，请谨慎判断。所有问题请用简体中简洁准确地回答。\end{CJK} \\[0.5ex]
\\[0.3em]

\textbf{English Version:} \\[0.5ex]
\textbf{System:} You are a helpful assistant. The documents I provide include documents that were originally written in \{\textcolor{blue}{query language}\} and others that are translations from English into \{\textcolor{blue}{query language}\} with quality evaluation scores. The translated documents are scored on semantic consistency, grammatical accuracy, and fluency, each ranging from 0.0 to 5.0. You should prioritize and rely on the original \{\textcolor{blue}{query language}\} documents first. Use the translated ones as sources in order of highest score. Be cautious with any uncertain information. Always answer as briefly and accurately as possible, and respond only in \{\textcolor{blue}{query language}\}. \\[0.5ex]
\noalign{\vskip 0.3ex}
\hline
\textbf{User Message} \\
\hline
Background: \{documents with quality scores\} \\[0.5ex]
Question: \{question\} \\
\hline
\end{tabular}
\caption{The prompt used for response generation. Documents with quality scores are provided to generator for better guidance. The system prompt explicitly instructs the model to prioritize higher-quality translations and respond only in query language.}
\label{tab:generation-prompt}
\end{table*}

\section{Translation Quality Assessment Cases}
\label{sec:appendix_translation_quality}

Tables \ref{tab:translation_quality_case_low} and \ref{tab:translation_quality_case_high} present case studies of our translation quality assessment process, demonstrating low- and high-quality translations. 

The cases in Table \ref{tab:translation_quality_case_low} show a translation with relatively low scores in all criteria. In Case 1: Korean,  it shows semantic distortions (e.g., "would not trigger a localized ice age" incorrectly translated as "would not occur"), grammatical errors including awkward sentence structures, and unnatural expressions that compromise fluency. In case 2: Finnish, it shows a translation with relatively low scores across all criteria (e.g., a quantity shift “to 10,000 Nazi war criminals” rendered as “over 10,000” and omissions of the El-Kurru and Nuri subsections) producing semantic distortions, grammatical issues, and reduced fluency. In Case 3: Chinese, it likewise shows low scores across all criteria (e.g., the title “Who Framed Roger Rabbit?” mistranslated as “Who fell into Roger’s trap”, “sense of humor” shifted to “original intention” and the proper name Dolores dropped to just “girlfriend” with duplicated tokens) leading to semantic drift, grammatical errors, and poor fluency.

The cases in Table \ref{tab:translation_quality_case_high} demonstrate high-quality translations across all three languages. 
In Case 1 (Korean), the output uses natural expressions and appropriate terminology (e.g., \begin{CJK}{UTF8}{mj}``주권''\end{CJK} for “states' rights”), accurately conveying complex political notions while maintaining readability. 
In Case 2 (Finnish), the translation preserves chronology and factual detail (e.g., correct date inflection “25.\ huhtikuuta 1945” and idiomatic phrasing such as “Kolmen valtakunnan rajapyykillä”), yielding strong grammatical accuracy and fluency. 
In Case 3 (Chinese), named entities and quantitative details are rendered precisely (e.g., \begin{CJK}{UTF8}{gbsn}“多用途体育场”、“可容纳8{,}000人”、“于2017年4月更名，以纪念格林纳达首位奥运奖牌得主基拉尼·詹姆斯”\end{CJK}), resulting in consistently high scores for semantic equivalence, grammatical accuracy, and fluency.

These cases illustrate how our quality assessment framework effectively captures the nuances of translation quality and provides meaningful metadata for the generation process.

\section{More cases of DKM-RAG and CrossRAG}
In Table~\ref{tab:dkm_hallucination_cases_finnish_chinese} and \ref{tab:crossrag_wrongtrans_finnish_chinese}, they show more failure cases in Finnish and Chinese queries. In DKM-RAG, during the refinement process, LLM tends to alter the content of the retrieved passage into query-related content, which distorts the actual meaning of the original retrieved passages. In CrossRAG, certain words are incorrectly translated by NLLB, which eventually leads the generator to rely on wrong passages. In both cases, these limitations result in failure to generate the correct answer.

\begin{table*}[t]
\centering\small
\begin{tabularx}{\columnwidth}{lX}
\toprule
\multicolumn{2}{l}{\textbf{Finnish}} \\
\midrule
\textbf{Query} & \textit{Mikä on Ilmestyskirja. Nyt -elokuvan genre?} (What is the genre of the movie Apocalypse Now?) \\
\textbf{Retrieved} & A war film directed by Francis Ford Coppola from 1979... \\
\textbf{Refined} & Francis Ford Coppolan ohjaama romanttinen komediaelokuva vuodelta 1979... (A romantic comedy film directed by Francis Ford Coppola from 1979...) \\
\textbf{LLM Output} & Romanttinen komediaelokuva. (Romantic comedy film.) \\
\midrule
\multicolumn{2}{l}{\textbf{Chinese}} \\
\midrule
\textbf{Query} & \begin{CJK}{UTF8}{gbsn}在我睡前要走几英里是谁写的 \end{CJK}(Who wrote "I walk a few miles before I go to sleep") \\
\textbf{Retrieved} & \begin{CJK}{UTF8}{gbsn}在我睡前要走几英里\end{CJK}(I have to walk a few miles before I sleep.)\\
\textbf{Refined} & \begin{CJK}{UTF8}{gbsn}我睡前写了几封信，还有几英里要走\end{CJK}(Before I sleep, I wrote a few letters and still had a few miles to go.) \\
\textbf{LLM Output} & \begin{CJK}{UTF8}{gbsn}这段话表示作者在睡眠前会同时完成两件\end{CJK}(This passage indicates that the author will complete two things at the same time before going to sleep.) \\
\bottomrule
\end{tabularx}
\caption{Case study of factual distortion in DKM-RAG for a Finnish and Chinese query.}
\label{tab:dkm_hallucination_cases_finnish_chinese}
\end{table*}

\begin{table*}[t]
\centering\small
\begin{tabularx}{\columnwidth}{lX}
\toprule
\multicolumn{2}{l}{\textbf{Finnish}} \\
\midrule
\textbf{Query} & \begin{CJK}{UTF8}{mj}kuka laulaa sugar pie honey bunch\end{CJK} (Who sings Sugar Pie Honey Bunch?") \\
\textbf{Retrieved} & I Can't Help Myself (Sugar Pie Honey Bunch). chart in 1989. The "Motorcity All-Stars" released a rerecorded version of song in 1989 on Ian Levine's Motown revival label Motorcity. \\
\textbf{Translated} & I Can't Help Myself (Sugar Pie Honey Bunch). sarja vuonna 1989. "Motorcity All-Stars" julkaisi uuden kappaleen vuonna 1989 Ian Levine'n Motown-elokuvalla Motorcity. (I Can't Help Myself (Sugar Pie Honey Bunch). series in 1989. "Motorcity All-Stars" released a new song in 1989 on Ian Levine's Motown movie Motorcity.) \\
\midrule
\multicolumn{2}{l}{\textbf{Chinese}} \\
\midrule
\textbf{Query} & \begin{CJK}{UTF8}{gbsn}'The world is like a great big onion' 是谁演唱的 \end{CJK} (Who sang "The world is like a great big onion"?)\\
\textbf{Retrieved} & The Onion Song — a 1969 hit by soul singers Marvin Gaye and Tammi Terrell. It entered the UK singles chart on 15 November 1969, peaking at No.9 on 9 December. Terrell died, aged 24, just prior to the U.S. single release. \\
\textbf{Translated} & \begin{CJK}{UTF8}{gbsn}洋歌曲。— “洋歌曲”是灵魂歌手马文·盖伊和塔米·特雷尔1969年的热门歌曲。这首歌在美国表现不佳。但它却成为盖伊和特雷尔在英国最热门的歌曲，于1969年11月15日进入单曲榜，并于12月9日最高排名第九。\end{CJK}("Western Song" was a 1969 hit by soul singers Marvin Gaye and Tammy Terrell. The song performed poorly in the United States. It became Gaye and Terrell's biggest hit in the United Kingdom, entering the singles chart on November 15, 1969, and peaking at No.9 on December 9.)\\
\bottomrule
\end{tabularx}
\caption{Case study of incorrect translation in CrossRAG for a Finnish and Chinese query.}
\label{tab:crossrag_wrongtrans_finnish_chinese}
\end{table*}

\begin{table*}[t]
\centering\small
\begin{tabularx}{\textwidth}{lX}
\toprule
\multicolumn{2}{l}{\textbf{Case 1: Korean}} \\
\midrule
\textbf{Original} & Retrieved English documents: The film implies that a geomagnetic pole-shift would trigger a localized ice age in Miami, although regions at lower latitudes receive more direct sunlight. A temperature drop to absolute zero (−273 °C) is scientifically impossible; before reaching −196 °C the two dominant atmospheric gases would liquefy and precipitate. \\
\textbf{Tagged} & \begin{CJK}{UTF8}{mj}영화는 암시하는 바와 같이 마이애미에 현지화된 빙하기가 발생하지 않을 것입니다. 지구 온도 감소 (최후 -273 °C) 를 경험하는 지구 지역의 묘사는 과학적으로 정확하지 않습니다. −196 °C (−320 °F) 아래는 지구 대기 중 두 가지 지배적인 가스가 액화되어 표면에 떨어질 것입니다. \colorbox{yellow}{[점수]} 의미론적 일치성: 2.5, 문법적 정확성: 2.0, 자연스러움과 유창성: 2.3\end{CJK} \\
\midrule
\multicolumn{2}{l}{\textbf{Case 2: Finnish}} \\
\midrule
\textbf{Original} & Egyptian pyramids. Egypt. Its builders reduced the amount of work necessary to construct it by ingeniously using as its foundation and core a 12-meter-high natural limestone hill.

Section: Number and location of pyramids.: El-Kurru.
Piye, the king of Kush who became the first ruler of the Twenty-fifth Dynasty of Egypt, built a pyramid at El-Kurru. He was the first Egyptian pharaoh to be buried in a pyramid in centuries.

Section: Number and location of pyramids.: Nuri.
Taharqa, a Kushite ruler of the Twenty-fifth Dynasty, built his pyramid at Nuri. It was the largest in the area (North Sudan) \\
\textbf{Tagged} & Egyptiläiset pyramidit. Egypti. Sen rakentajat vähensivät sen rakentamiseen tarvittavan työn määrää käyttämällä älykkäästi pohjana ja ytimenä 12 metrin korkean luonnollisen kalkkikivikukkulan. \colorbox{yellow}{[pisteet]} Semanttinen johdonmukaisuus: 2.5, Kieliopillinen tarkkuus: 2.0, Luontevuus ja sujuvuus: 2.0 \\
\midrule
\multicolumn{2}{l}{\textbf{Case 3: Chinese}} \\
\midrule
\textbf{Original} & Who Framed Roger Rabbit? Eddie Valiant, having fully regained his sense of humor, begins moving forward with his girlfriend, Dolores. The residents of Toontown gather to celebrate the victory. Bob Hoskins portrays Eddie Valiant, a private investigator with a soft spot for toons and a weakness for drink. \\
\textbf{Tagged} & \begin{CJK}{UTF8}{gbsn}谁陷入子罗杰的陷.自完全全重拾初心 ,与女友友友迈步前行 ,动画 城众人也开 始庆祝他人的胜利.演员勃勃 · 霍斯金斯饰 艾迪·瓦利安特 , 一位对动画角色心有喜欢的空隙而又 酒的私家侦探. \colorbox{yellow}{[分数]} 语义一致性: 2.5, 语法准确性: 1.0, 语言流畅度: 1.0\end{CJK} \\
\bottomrule
\end{tabularx}
\caption{Case studies of low-quality translation assessment in Korean, Finnish and Chinese. Quality scores are provided for semantic equivalence, grammatical accuracy, and fluency}
\label{tab:translation_quality_case_low}
\end{table*}

\begin{table*}[t]
\centering\small
\begin{tabularx}{\textwidth}{lX}
\toprule
\multicolumn{2}{l}{\textbf{Case 1: Korean}} \\
\midrule
\textbf{Original} & John Tyler (March 29, 1790 ~ January 18, 1862) was the tenth president of the United States, serving from 1841 to 1845, after briefly holding office as the tenth vice president in 1841. He was elected vice president on the 1840 Whig ticket with President William Henry Harrison, succeeding to the presidency following Harrison's death 31 days after assuming office. Tyler was a stalwart supporter and advocate of states' rights, and he adopted nationalistic policies as president only when they did not infringe on the states' powers. \\
\textbf{Tagged} & \begin{CJK}{UTF8}{mj}존 타일러 (1790년 3월 29일 ~ 1862년 1월 18일) 는 1841년부터 1845년까지 10대 부통령을 지낸 후 10대 대통령이다. 그는 1840년 윌리엄 헨리 해리슨 대통령과 함께 위기당 당선되었다. 타일러는 1841년 4월 해리슨의 사망 후 대통령직에 올랐다. 새로운 행정부 출범 후 불과 한 달 만에 대통령직에 올랐다. 그는 주권 지지자이자 주권 옹호자였으며, 대통령으로서 국가주의 정책을 채택했을 때만 주권권을 침해하지 않았다. \colorbox{yellow}{[점수]} 의미론적 일치성: 4.5, 문법적 정확성: 4.8, 자연스러움과 유창성: 4.5\end{CJK} \\
\midrule
\multicolumn{2}{l}{\textbf{Case 2: Finnish}} \\
\midrule
\textbf{Original} & End of World War II in Europe. to 10,000 Nazi war criminals eventually fled Europe using ratlines such as ODESSA. German forces leave Finland: On 25 April 1945, the last German troops withdrew from Finnish Lapland and made their way into occupied Norway. On 27 April 1945, the Raising the Flag on the Three-Country Cairn photograph was taken. Mussolini's death: On 25 April 1945, Italian partisans liberated Milan and Turin. On 27 April 1945, as Allied forces closed in on Milan, Italian dictator Benito Mussolini was captured by Italian partisans. It is disputed whether he was trying to flee from Italy to Switzerland (through the Splügen Pass). \\
\textbf{Tagged} & Toisen maailmansodan lopussa Euroopassa yli 10 000 natsisotarikollista pakeni mantereelta ODESSA:n kaltaisten ratline-reittien kautta. 
Saksalaiset joukot poistuivat Suomesta: 25. huhtikuuta 1945 viimeiset saksalaiset vetäytyivät Lapista miehitettyyn Norjaan. 
27. huhtikuuta otettiin valokuva, jossa nostettiin lippu Kolmen valtakunnan rajapyykillä. 
Mussolinin kuolema: 25. huhtikuuta italialaiset partisaanit vapauttivat Milanon ja Turinin. 
27. huhtikuuta, kun liittoutuneet lähestyivät Milanoa, Benito Mussolini vangittiin partisaanien toimesta.
\colorbox{yellow}{[pisteet]} Semanttinen johdonmukaisuus: 4.5, Kieliopillinen tarkkuus: 4.5, Luontevuus ja sujuvuus: 4.5 \\
\midrule
\multicolumn{2}{l}{\textbf{Case 3: Chinese}} \\
\midrule
\textbf{Original} & Kirani James Athletic Stadium is a multi-purpose stadium in St.\ George’s, Grenada. Formerly known as the Grenada National Stadium, it is used mostly for football matches and has a capacity of 8,000. In April 2017 the venue was renamed Kirani James Athletic Stadium in honour of Kirani James, Grenada’s first Olympic medallist. \\
\textbf{Tagged} & \begin{CJK}{UTF8}{gbsn}基拉尼詹姆斯竞技体育场警察场是格莱纳达圣乔治的多用途体育场.目前主要用于足球比赛.该体育场容纳8000人.于2017年4月改名为基拉尼詹姆斯竞技体育场,以纪念格莱纳达第一个奥运奖得主基拉尼詹姆斯. \colorbox{yellow}{[分数]} 语义一致性: 4.8, 语法准确性: 4.5, 语言流畅度: 4.2\end{CJK} \\
\bottomrule
\end{tabularx}
\caption{Case studies of high-quality translation assessment in Korean, Finnish, and Chinese. Quality scores are provided for semantic equivalence, grammatical accuracy, and fluency.}
\label{tab:translation_quality_case_high}
\end{table*}

\end{document}